\documentclass[letterpaper, 10 pt, conference]{ieeeconf} 
\pdfoutput=1
\IEEEoverridecommandlockouts  
\usepackage{graphics} 
\usepackage{epsfig}
\usepackage{mathptmx} 
\usepackage{times}
\usepackage{amsmath} 
\usepackage{amssymb} 
\usepackage{url}
\usepackage{color}
\usepackage{graphicx}
\usepackage{subcaption}
\usepackage{caption}
\usepackage{multirow}
\usepackage{url}
\usepackage{float}
\usepackage{array}
\usepackage{relsize}
\usepackage{sidecap}   
\usepackage{mwe} 
\usepackage{cite}
\usepackage{tikz}
\usepackage{textcomp}

\newcommand\copyrighttext{
	\footnotesize \textcopyright 2023 IEEE. Personal use of this material is permitted.  Permission from IEEE must be obtained for all other uses, in any current or future media, including reprinting/republishing this material for advertising or promotional purposes, creating new collective works, for resale or redistribution to servers or lists, or reuse of any copyrighted component of this work in other works.
	DOI: 10.1109/ITSC57777.2023.10422417}
\newcommand\copyrightnotice{
	\begin{tikzpicture}[remember picture,overlay]
		\node[anchor=south,yshift=10pt] at (current page.south) {\fbox{\parbox{\dimexpr\textwidth-\fboxsep-\fboxrule\relax}{\copyrighttext}}};
	\end{tikzpicture}
}

\title{\LARGE \bf
Improved context-sensitive transformer model for inland vessel trajectory prediction
}

\author{Kathrin Donandt$^{1}$, Karim Böttger$^{2}$, and Dirk Söffker$^{3}$
\thanks{$^{1}$Kathrin Donandt is with the Institute of Ship Technology, Ocean Engineering and Transport Systems (ISMT), University of Duisburg-Essen, Duisburg, Germany. 
        {\tt\small kathrin.donandt@uni-due.de}}
\thanks{$^{2}$Karim Böttger is with the German Federal Waterways Engineering and Research Institute, Karlsruhe, Germany.  
	{\tt\small karim.boettger@baw.de}}
\thanks{$^{3}$Dirk Söffker is with the Chair of Dynamics and Control (SRS), University of Duisburg-Essen, Duisburg, Germany. 
        {\tt\small soeffker@uni-due.de}}
}

\begin{document}

\maketitle
\copyrightnotice
\thispagestyle{empty}
\pagestyle{empty}

\begin{abstract}
Physics-related and model-based vessel trajectory prediction is highly accurate but requires specific knowledge of the vessel under consideration which is not always practical. Machine learning-based trajectory prediction models do not require expert knowledge, but rely on the implicit knowledge extracted from massive amounts of data. Several deep learning (DL) methods for vessel trajectory prediction have recently been suggested. 
The DL models developed typically only process information about the (dis)location of vessels defined with respect to a global reference system. In the context of inland navigation, this can be problematic, since without knowledge of the limited navigable space, irrealistic trajectories are likely to be determined. If spatial constraintes are introduced, e.g., by implementing an additional submodule to process map data, however, overall complexity increases. Instead of processing the vessel displacement information on the one hand and the spatial information on the other hand, the paper proposes the merging of both information. Here, fairway-related and navigation-related displacement information are used directly. In this way, the previously proposed context-sensitive Classification Transformer (CSCT) shows an improved spatial awareness. Additionally, the CSCT is adapted to assess the model uncertainty by enabling dropout during inference. 
This approach is trained on different inland waterways to analyze its generalizability. As the improved CSCT obtains lower prediction errors and enables to estimate the trustworthiness of each prediction, it is more suitable for safety-critical applications in inland navigation than previously developed models.
 
\end{abstract}


\begin{figure}
	\begin{subfigure}{.24\textwidth}
		\includegraphics[width=\linewidth]{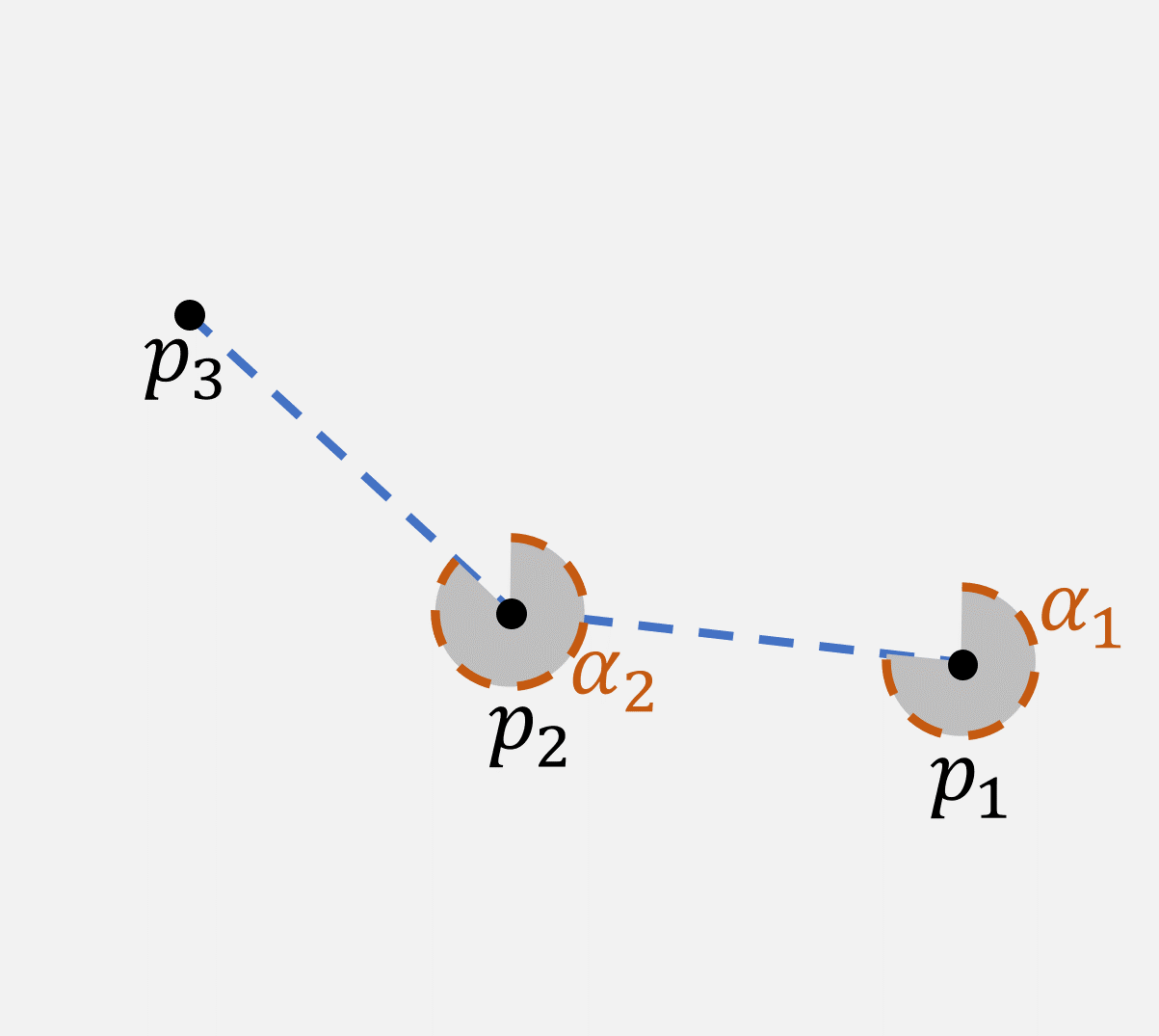}
		\caption{Global RS}
		\label{fig:ref_sys_glob}
	\end{subfigure}
	\begin{subfigure}{.24\textwidth}
		\includegraphics[width=\linewidth]{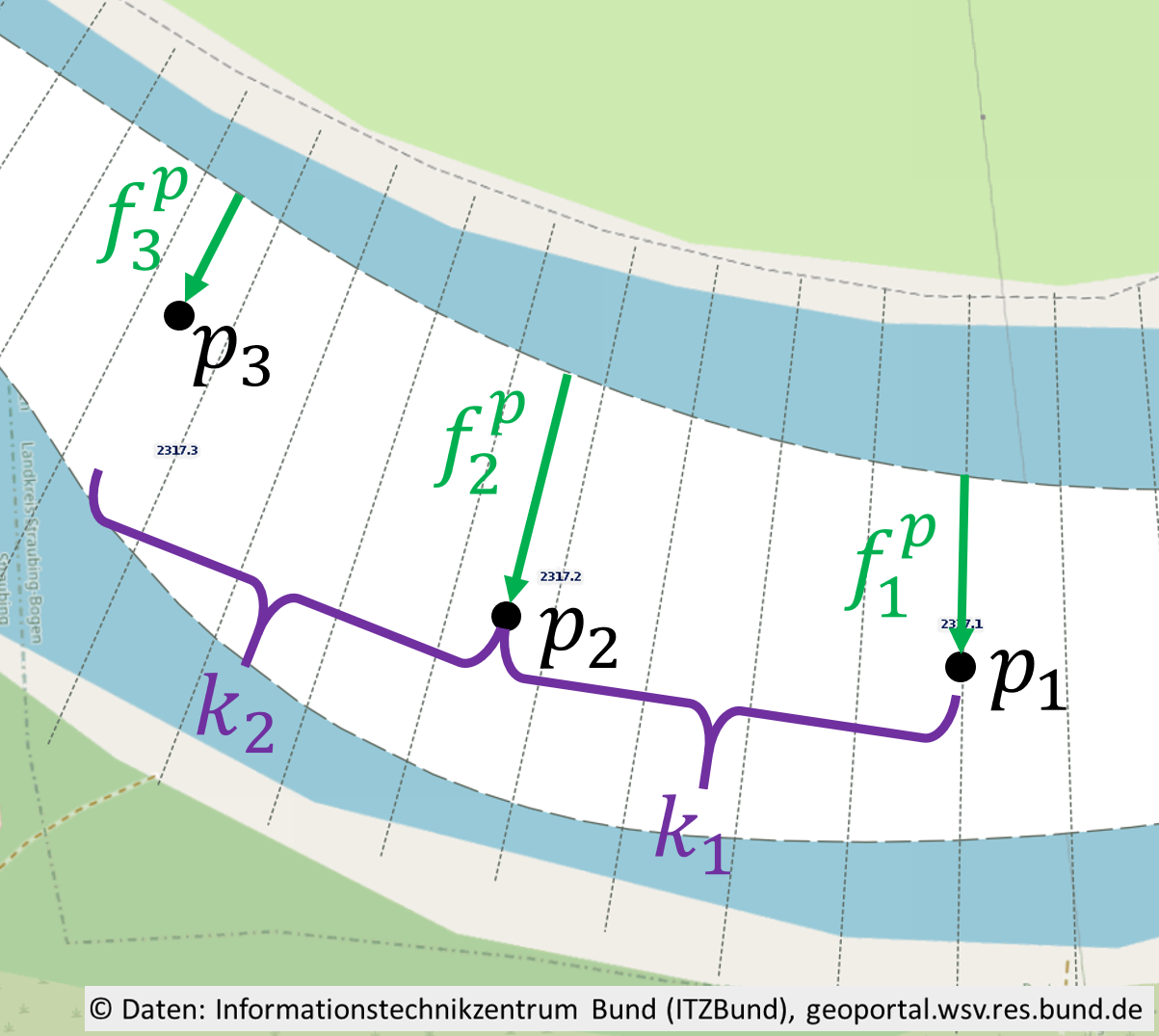}
		\caption{River-specific RS}
		\label{fig:ref_sys_riv}
	\end{subfigure}\par
	\begin{subfigure}{.24\textwidth}
		\includegraphics[width=\linewidth]{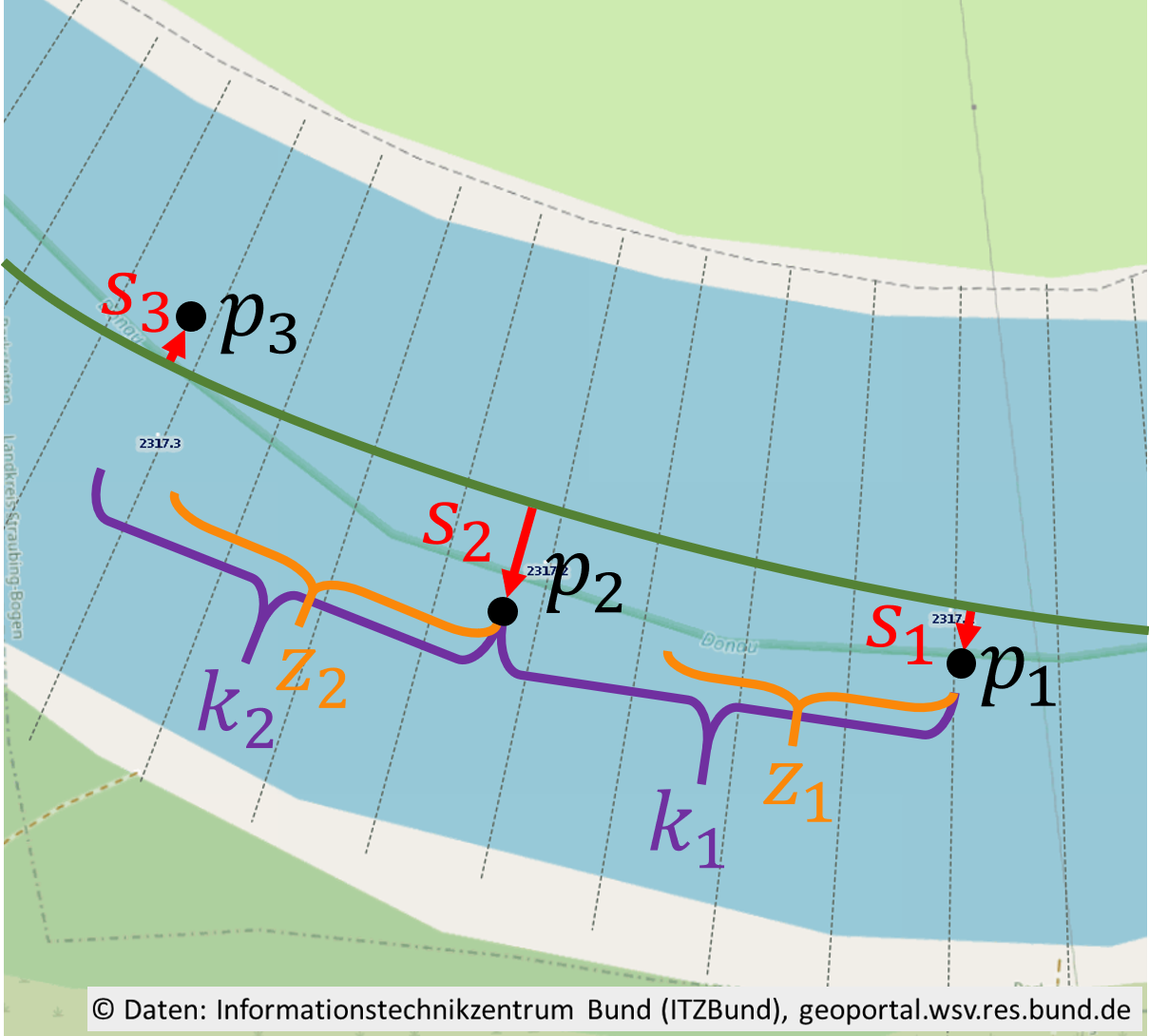}
		\caption{Navigation-specific RS}
		\label{fig:ref_sys_nav}
	\end{subfigure}
	\begin{subfigure}{.24\textwidth}
		\includegraphics[width=\linewidth]{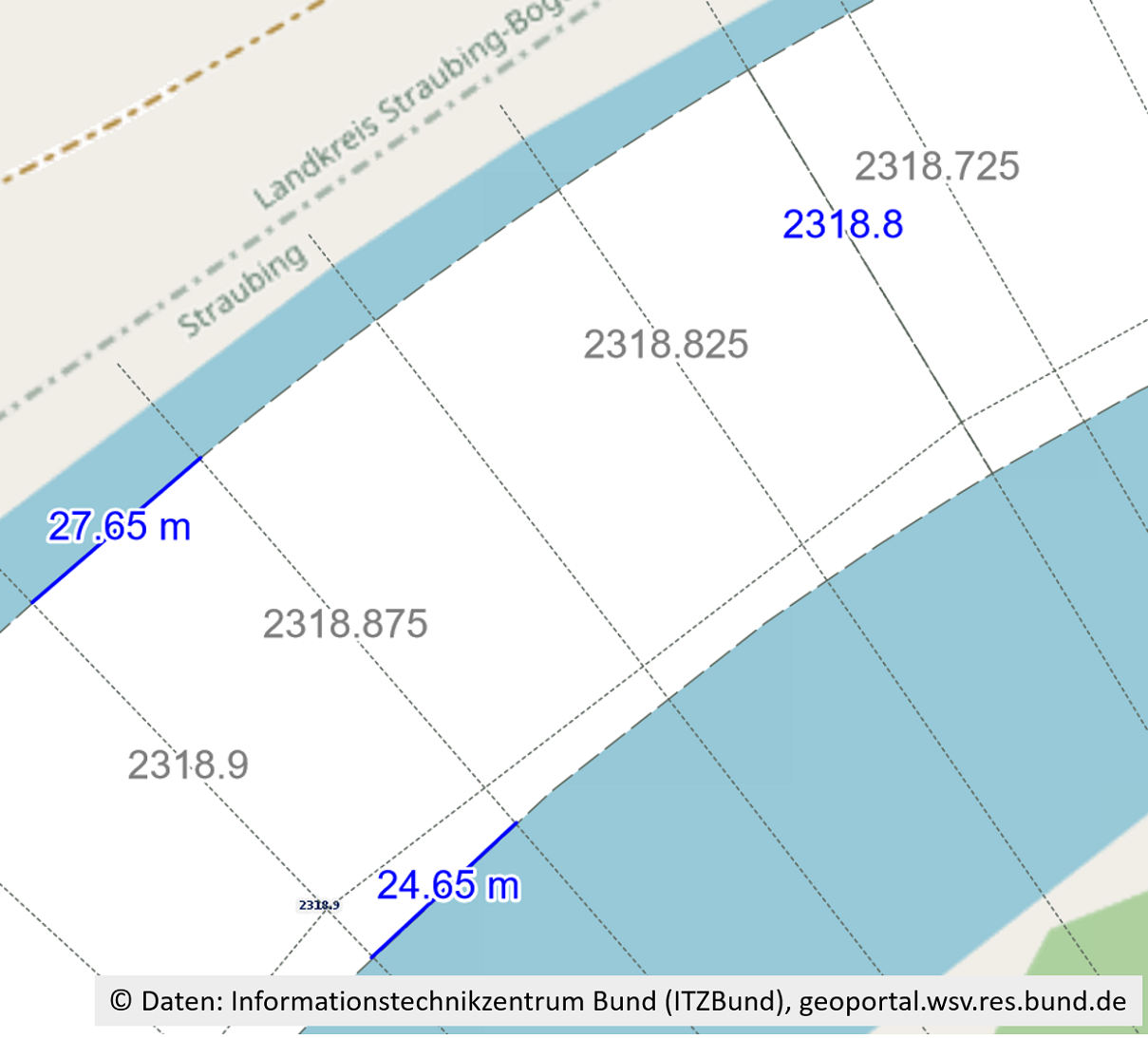}
		\caption{Non-equidistant KM}
		\label{fig:wwkmjump}
	\end{subfigure}
			\caption{Reference systems (RS) for dislocation information (a) - (c) and non-equidistant Danube kilometers (KM) (d)
			}
		 \vspace{-1.7em}
		\end{figure}
\section{INTRODUCTION}
Automation of ship guidance is a current and urgent problem, as the ship is considered the most important and environmentally friendly means of transport in global trade and the shipping industry suffers from a shortage of skippers \cite{Negenborn.2023}. Automatic vessel trajectory prediction (VTP) is a fundamental, safety-critical task required in this endeavor, as it supports collision avoidance, route planning, and anomaly detection applications \cite{Zhang.2022}. 
The research on VTP is increasingly making use of deep learning (DL) methods. In a recent review on the subject \cite{Zhang.2022} a growth of the share of publications reporting the usage of DL methods in the overall VTP literature from 38 \% in 2018 to 77 \% in 2021 is mentioned. Most approaches are developed for maritime vessels and use vessel movement data extracted from Automatic Identification System (AIS) logs \cite{Zhang.2022}. Especially in the case of inland VTP, these data alone are insufficient for realistic predictions as the limitation of the navigable area is unknown. Few approaches include information on the spatial constraints by making use of further data sources such as radar images, electronic navigational charts \cite{DijtPimandMettesPascal.2020}, or shoreline data \cite{Mehri.2021}. 
Instead of realizing spatial awareness by increasing the numbers of features and the complexity of the DL model through an additional map-processing sub-module, dislocation and navigation space information are merged. This is realized by introducing alternatives to the global or ship-centered reference systems for vessel movement representation in which the lateral and longitudinal dislocation are defined in relation to the fairway or to typical navigation patterns extracted from historical data. 
A further marginally addressed aspect in the research on DL-based VTP is the consideration of uncertainty. 
In safety-critical applications, it is important to be able to know when a DL model is incapable of producing reliable predictions. 
Two types of uncertainty, aleatoric and epistemic, are often distinguished, where the former refers to the irreducible uncertainty given by the data itself, and the latter to the model (parameter) uncertainty due to the limited knowledge the model is able to extract from the provided training data \cite{Abdar.2021,Kendall.2017}. In this contribution, the Monte Carlo (MC) dropout method for the epistemic uncertainty quantification of the previously proposed context-sensitive classification transformer (CSCT) \cite{Donandt.2022} is applied. The epistemic uncertainty evolution over the prediction horizon and the relationship between the prediction error and uncertainty are analyzed. Finally, the CSCT variants are trained and evaluated on data of the Danube and Rhine river to prove the generalizability of the proposed approach. 

\section{RELATED WORK}\label{sec:RelatedWork} 
Deep learning-based inland vessel trajectory prediction approaches commonly use a time series of vessel positions given in latitude and longitude coordinates (e.g. ~ref\cite{Feng.2022,Yuan.2020,Jiang.2021,Volkova.2021,You.2020,Mehri.2021}), the distances between such coordinates (e.g.\cite{Gao.2021}),  speed-over-ground and/or course-over-ground (COG) information (e.g. \cite{Gao.2021, You.2020,Mehri.2021}). Sometimes, a ship-specific coordinate system is used, where the origin is at the vessel's last known position and the orientation is given by the last known COG (e.g. \cite{DijtPimandMettesPascal.2020}). These (dis)location features of the time series characterizing the vessel's movement are problematic in restricted waterways as the vessels cannot navigate in the whole area of the considered coordinate system. 
Thus, the information of where the vessels can navigate needs to be additionally considered, as e.g. done in \cite{DijtPimandMettesPascal.2020} by the usage of radar images and electronic navigational charts. The usage of map data to inform a DL-based trajectory prediction model about the navigable area is also common in the automotive domain (see e.g. the consideration of ``road-related factors'' in many approaches presented in \cite{Huang.2022}). 

Despite the relevance of uncertainty consideration in the context of safety-critical applications, the epistemic and aleatoric uncertainties involved in the development of DL-based VTP models are rarely taken into account. 
In \cite{Capobianco.2021} and \cite{Liu.2022}, the epistemic uncertainty of offline trained models is quantified through the usage of Monte Carlo (MC) dropout during inference. In \cite{Gao.2021}, the epistemic uncertainty of a model trained online  before each prediction is approximated by the confidence interval of the training errors assumed to have a Gaussian distribution. The aleatoric uncertainty is considered in \cite{KristianAallingSrensen.2022} and in addition to the epistemic uncertainty in \cite{Capobianco.2021}. 
A model predicting the parameters of a probability density function from which the next probabilistic locations can be obtained is proposed in both publications. 

\section{IMPROVED DISLOCATION DATA FOR TRAJECTORY PREDICTION}\label{sec:ImprovedDislocationData}
\begin{figure}
	\centering
	\begin{subfigure}{0.49\textwidth}
		\centering
		\includegraphics[width=\textwidth]{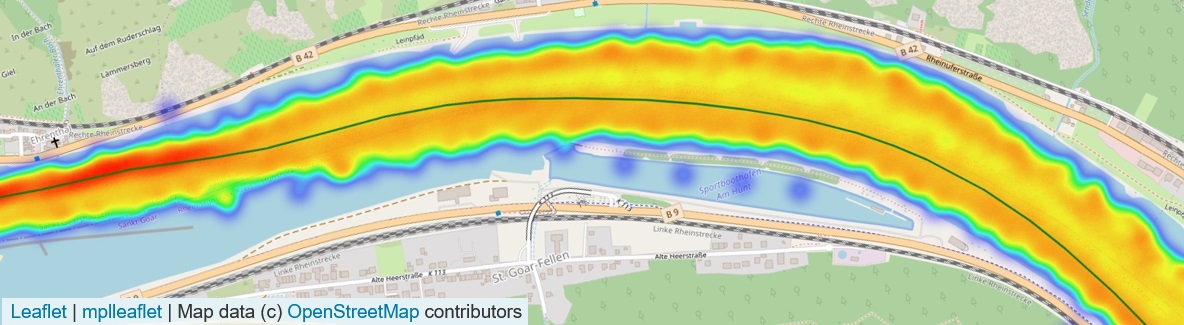}
		\caption{}\label{fig:trrhein}
	\end{subfigure}
	\vfill
	\begin{subfigure}{0.49\textwidth}
		\centering
		\includegraphics[width=\textwidth]{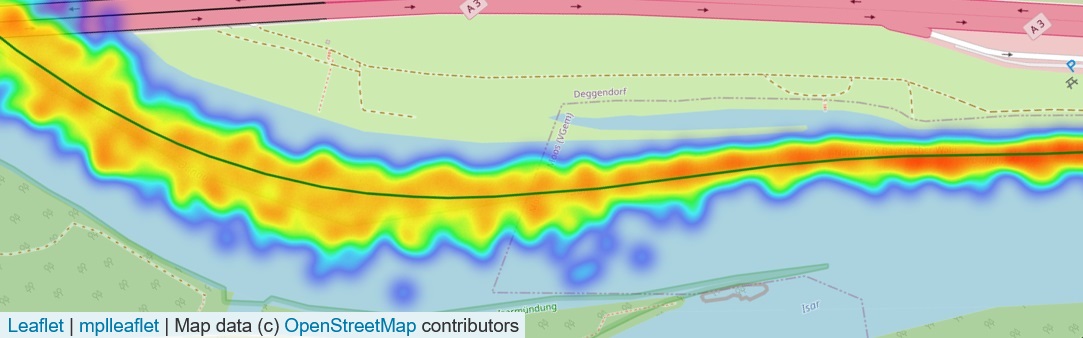}
		\caption{}\label{fig:trdonau}
	\end{subfigure}
\begin{subfigure}{0.49\textwidth}
	\centering
	\includegraphics[width=\textwidth]{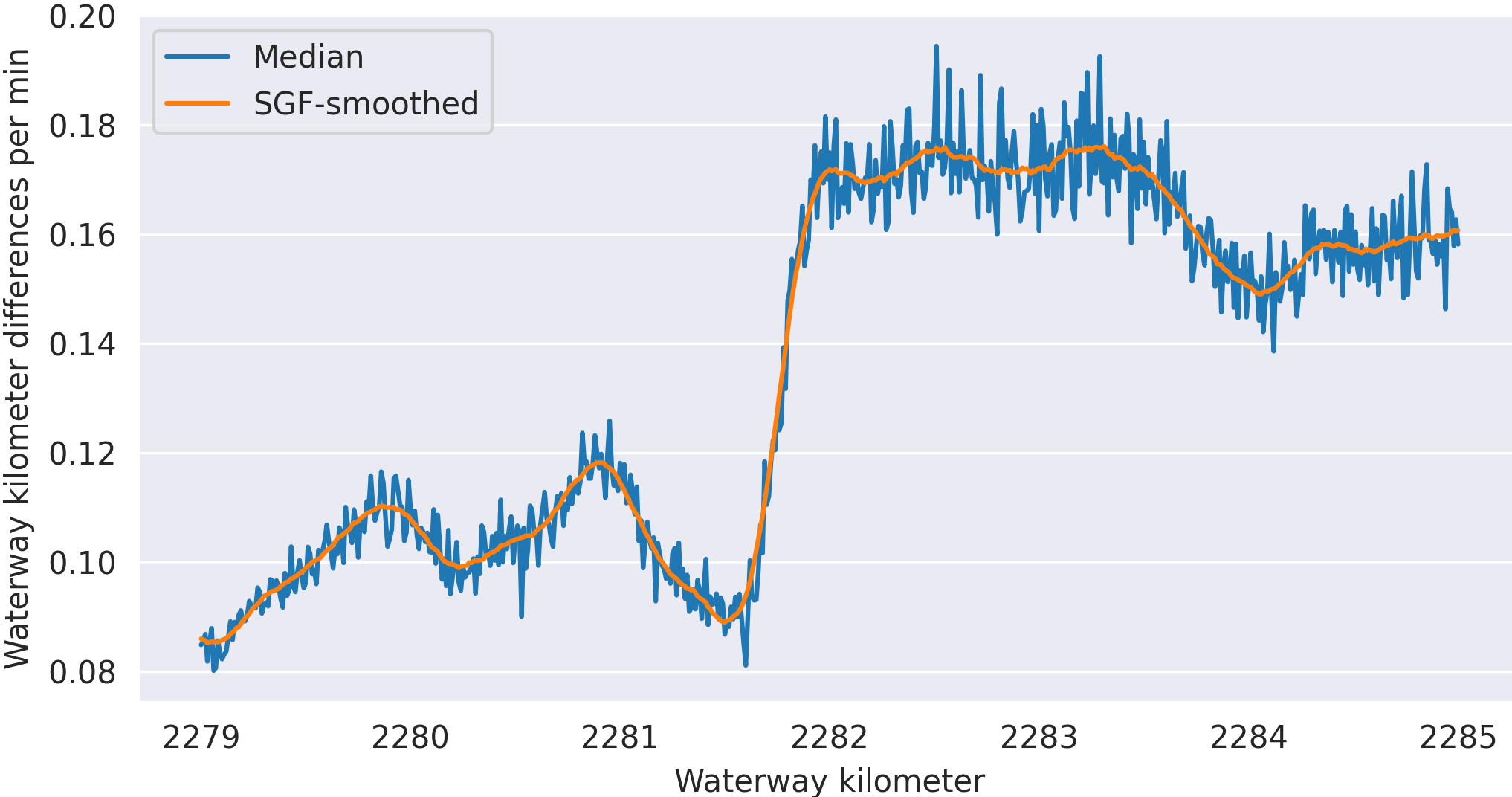}
	\centering
	\caption{}\label{fig:savgol_speeds}
\end{subfigure}
\caption{N-CSCT reference system: (a) typical route (green line) and AIS data (heatmap), (b) typical KM distances per minute (orange) and underlying medians from AIS data}
\vspace{-1.5em}
\end{figure} 
The features of time series processed by DL-based inland vessel trajectory prediction models which describe the vessel motion consist of at least one lateral and one longitudinal (dislocation) component related to a global or ship-specific reference system and are referred to as $x_t^{glob} \in \mathbb{R}^2$, where $t$ is the time step index. For example, $x_t^{glob}$ of the previously suggested context-sensitive classification transformer model \cite{Donandt.2022}, referred to as G-CSCT here, is defined as 
\begin{equation}
	x_t^{glob} = \Bigl( \lVert p_{t+1}-p_t \rVert, \alpha_{t+1}-\alpha_t\Bigr),
	\label{formula:xtg}
\end{equation}
where $p_t$ refers to the vessel's position in UTM coordinates and $\alpha_t$ to the COG angle at time step $t$. 
The longitudinal and lateral components therefore consist of the Euclidean distance and the COG change between subsequent vessel positions, respectively (Fig. \ref{fig:ref_sys_glob}). 
The G-CSCT model is unaware of the spatial limitations given by the navigable area. Instead of providing additional input features describing the spatial constraints, the reference system of the lateral and longitudinal dislocation information can be modified. Here, a river-specific reference system based on the river's subdivision into hectometer sections and on the fairway is proposed. 
Vessels typically navigate on the fairway where a sufficient water depth is guaranteed. 
The model trained on river-referenced data, R-CSCT, processes time series with the features $x_t^{riv}$. The longitudinal component of $x_t^{riv}$ consists of the distance between the waterway kilometers (KMs) of subsequent vessel positions. 
The KMs are based on the hectometers and profile lines defined for (German) waterways. For a requested geographical position, waterway ID, KM, distance from and position in relation to the river axis are obtained by an algorithm here denoted as kilometerization. Therefore the nearest river axis profiles are searched by a balanced, k-dimensional binary search tree. The enclosing profiles and kilometerization are then determined by geometry and interpolation. 
The function $h$ maps any coordinate to the corresponding KM using this kilometerization algorithm. Points on a single river profile line (dotted gray lines in Figs. \ref{fig:ref_sys_riv} and \ref{fig:ref_sys_nav}) belong to one and the same KM. Thus, the Euclidean distance between two profile lines naturally vary along the river width in curves, whereas the KM distance stays the same (see Fig. \ref{fig:wwkmjump}). 
Some rare exceptions exist where the Euclidean and the KM distances do not differ due to the non-parallelism of the profile lines but as a consequence of waterway kilometer gaps (see also Fig. \ref{fig:wwkmjump}) which are historically given or an outcome of river engineering measures.  
The problem of non-equidistance between waterway kilometers in curves is mitigated by considering curvature values as suggested in \cite{Donandt.2022} and additionally the curve orientation (in navigation direction perspective). The waterway kilometer gaps are addressed by internally processing shifted waterway kilometers instead of the official ones. The lateral dislocation component of $x_t^{riv}$ is defined by the change of the vessel's relative distance to the right fairway boundary (in navigation direction perspective), as skippers are recommended to follow a right-hand traffic whenever this is possible. The distance to the fairway boundary of a vessel at position $p_t$ can be approximated by the Euclidean distance between $p_t$ and the coordinate on the fairway boundary at KM $h(p_t)$. To retrieve a fairway boundary coordinate for any given KM, a second order spline interpolation is applied to fairway boundary coordinates provided by the German Federal Waterways and Engineering Institute (BAW). This yields the function $r$ for the right and $l$ for the left fairway boundary. 
Formally, $x_t^{riv}$ is given by  
\begin{equation}
	x_t^{riv} = \biggl(k_t, \frac{f^p_{t+1}}{w_{t+1}}-\frac{f^p_t}{w_t}\biggr),
	\label{formula:riverSpec}
\end{equation}
with 
\begin{equation*}
	k_t = h(p_{t+1}) - h(p_t)
\end{equation*}
\begin{equation*}
	w_t = \lVert r(h(p_t))-l(h(p_t)) \rVert
\end{equation*}
\begin{equation*}
	f_t^p = \begin{cases}
		- d_t^{rp}, & \text{if $w_t<d_t^{lp}$ and $d_t^{rp}<d_t^{lp}$ }\\
		+ d_t^{rp} & \text{otherwise},
	\end{cases}
\end{equation*}
where
\begin{equation*}
	d_t^{rp} = \lVert r(h(p_t))-p_t \rVert,
\end{equation*}
and $d_t^{lp}$ is defined in analogy to $d_t^{rp}$ by replacing $l$ with $r$. Thus, $f_t$ is positive (negative) whenever the vessel is located left (right) of the right fairway boundary. In Fig. \ref{fig:ref_sys_riv}, the meaning of $f_t^p$ and the longitudinal component $k_t$ are visualized. The white area depicts the fairway.  
Instead of the river-specific reference system, the vessel dislocation can also be defined in relation to a typical route and typical KM distances travelled per minute, which can both be obtained from a statistical evaluation of historical AIS data. This is referred to as navigation-specific reference system and the corresponding CSCT model is denominated N-CSCT.  
A typical route is a more expressive reference line than the fairway boundary as it implies the knowledge of where the vessels typically navigate w.r.t. the lateral extension of the river.  
Deviations from typical KM distances per minute are advantageous over $k_t$ as region-specific speed variations due to flow acceleration caused by a river estuary or speed limits are taken into account. 
The lateral component of the time series feature $x_t^{nav}$ of N-CSCT is given by the change of the vessel's distance to the direction-specific typical route schematically depicted in green in Fig. \ref{fig:ref_sys_nav}. This route is determined by applying a Savitzky-Golay-Filter (SGF) on the easting and northing values from direction-specific historical AIS data, sorted and grouped by waterway decameters.  
A function $q$ is obtained that returns for any given KM the corresponding position on the typical route.  
Examples of upstream typical route sections of the Rhine and Danube river are given in Fig. \ref{fig:trrhein} and \ref{fig:trdonau} in green, respectively. The AIS data is spread over the whole navigable Rhine width in the curve, but the typical route indicates that the vessels prefer to navigate on the inner curve, which is characteristic for upstream navigation. 
The longitudinal component of $x_t^{nav}$ expresses the deviation from the direction-specific typical KM distance per minute. The function $z$, which returns for a given KM this typical KM distance per minute, is obtained by applying a SGF over medians of direction-specific historical KM distances per minute grouped by waterway decameter.   
In Fig. \ref{fig:savgol_speeds}, $z$ is denoted in orange for a Danube section (upstream direction). The jump at 2281.7 is due to the estuary of the Isar river. 
The formal definition of $x_t^{nav}$ is given by
\begin{equation}
	x_t^{nav} = \Bigl( k_t-z(h(p_t)), s_{t+1}-s_t \Bigr),
\end{equation} 
with
\begin{equation*}
	s_t = \begin{cases}
		+ d_t^{sp}, &  \text{if $f_t^p>f_t^{nav}$} \\ 
		- d_t^{sp} & \text{otherwise}
	\end{cases}
\end{equation*}
where
\begin{equation*}
	d_t^{sp} = \lVert q(h(p_t))-p_t \rVert.
\end{equation*}
The approximate distance of the typical route from the fairway boundary at KM $h(p_t)$, $f_t^{nav}$, can be obtained by replacing $p_t$ with $q(h(p_t))$ in the formula for $d_t^{rp}$ in Eq. \ref{formula:riverSpec}. In Fig. \ref{fig:ref_sys_nav}, the meaning of the dislocation calculation is depicted, where $\text{z}_t$ denotes $z(h(p_t))$. For N-CSCT, an improved consideration of the non-equidistance of waterway hectometers compared to R-CSCT is realized. The Euclidean distance between two positions of a KM distance of 0.1 on the typical route is additionally passed to the N-CSCT together with the curvature value and curve orientation. 
To illustrate the non-equidistance problematic in the case of N-CSCT, the case of $k_t-z(h(p_t)) = 0$ is considered. A vessel navigating on a straight river section is then travelling with the same speed as typically observed there independently of its distance from the typical route. 
In a curve, it does not become clear if the vessel is in fact moving with the typical waterway kilometer distance per minute or not. If the typical route lies on the outer curve, and the vessel passes on the inner, it has a lower speed. In the inverted case, it moves faster than the majority. The model can distinguish between these cases if, in addition to the vessel's position in relation to the typical route, information on the curve orientation and the position of the typical route in curves is available. For the latter, a 
Euclidean distance between two waterway hectometers that is numerically higher than the KM distance indicates that the typical route lies on the outer curve.

\section{UNCERTAINTY-AWARE CONTEXT-SENSITIVE TRANSFORMER MODEL}
Deep learning models typically do not permit insights into prediction confidence. Especially in safety-critical applications this is problematic. Information about the degree of uncertainty for a given prediction could prevent wrong decisions based on too uncertain predictions. 
Thus, apart from the modified input feature definition, the previously proposed CSCT model is transformed into a uncertainty-aware model. 
The epistemic uncertainty is quantified through the usage of MC dropout during inference as proposed in \cite{Gal.2016}. 
The dropout mechanism has originally been proposed to prevent overfitting in neural networks by randomly dropping neurons during training \cite{Nitish.2014}. By repeating the forward pass through a neural network with activated dropout at test time, different results are obtained for the same input. The larger the standard deviation of these results, the less confident is the model about its prediction for the given input.   
By returning the uncertainty degree together with the prediction of the CSCT, it will become possible to determine the necessity of a fallback solution to prevent collision avoidance strategies that are based on highly uncertaint trajectory predictions of surrounding vessels. 

\section{RESULTS}  
\subsection{Data}
The data preprocessing is done similar to \cite{Donandt.2022}. The previous linear interpolation between the irregularly sampled AIS logs is replaced by a cubic Hermite spline interpolation to obtain smoother results. The observation time $T_{obs}$ and prediction window $n$ are set to 5 and 10, respectively, and the time step size between subsequent observations/predictions is set to 1 min. The discretization of the dislocation features required due to the classification reframing of the regression problem (see \cite{Donandt.2022} for details) causes unavoidable imprecisions. Higher resolutions can decrease this imprecision, but can lead to a number of classes too high for the model to effectively learn on the given amount of data. 
The resolution of dislocation features given in meter or KM is set to 1 m / 0.001 KM in analogy to the original CSCT. The remaining resolutions are chosen empirically. 
An overview of the utilized resolutions and the corresponding average discretization errors between original coordinates and coordinates reconstructed from discretized data after 10 time steps is given in Table \ref{tab:discrError}. The  imprecisions are low even for R-CSCT and N-CSCT which rely on the non-equidistant waterway hectometers in their dislocation feature definitions. 
\begin{table}[]	
	\centering
	{\setlength\extrarowheight{2pt}
		\caption{Discretization resolutions and average discretization errors (Rhine dataset)}\label{tab:discrError}
		\resizebox{\columnwidth}{!}{%
			\begin{tabular}{c|c|c|c|}
				\cline{2-4}
				& \textbf{G-CSCT} & \textbf{R-CSCT}  & \textbf{N-CSCT}  \\ \hline
				\multicolumn{1}{|c|}{\textbf{\begin{tabular}[c]{@{}c@{}}Longitudinal resolution\end{tabular}}} & 1 m             & 0.001 waterway km & 0.001 waterway km \\ \hline
				\multicolumn{1}{|c|}{\textbf{\begin{tabular}[c]{@{}c@{}}Lateral resolution\end{tabular}}}      & 0.5 °           & 0.005            & 1 m              \\ \hline
				\multicolumn{1}{|c|}{\textbf{\begin{tabular}[c]{@{}c@{}}Discretization error\end{tabular}}}    & 5.15 m          & 1.00 m           & 1.19 m           \\ \hline
		\end{tabular}}
	}
\vspace{-0.5em}
\end{table}
The temporal and geographic range of the AIS data and the sizes of the extracted training datasets (i.e. the number of sequences) are given in Table \ref{tab:data}. 
The smaller size of the Danube dataset is owed to the lower traffic volumes compared to the Rhine. Direction-specific models are developed, and for this study, the upstream direction is considered. The models are trained separately on each dataset. 

\begin{table}[]
	\centering
	{\setlength\extrarowheight{2pt}
		\caption{Dataset specification}\label{tab:data}\resizebox{\columnwidth}{!}{%
		\begin{tabular}{c|c|c|}
			\cline{2-3}
			& \textbf{Danube dataset} & \textbf{Rhine dataset}                                                     \\ \hline
			\multicolumn{1}{|c|}{\textbf{Temporal range}}                                                                         & 330 days           & 328 days                                                                \\ \hline
			\multicolumn{1}{|c|}{\textbf{\begin{tabular}[c]{@{}c@{}}Spatial range (waterway km)\end{tabular}}}         & 2231-2321          & \begin{tabular}[c]{@{}c@{}}556-562, 570-579,598-608\end{tabular} \\ \hline
			\multicolumn{1}{|c|}{\textbf{\begin{tabular}[c]{@{}c@{}}Training dataset size\end{tabular}}} & 37k & 66k \\ \hline			\multicolumn{1}{|c|}{\textbf{\begin{tabular}[c]{@{}c@{}}Test dataset size\end{tabular}}} & 4.9k & 9.6k \\ \hline
	\end{tabular}}}
\vspace{-1.2em}
\end{table}
\subsection{Implementation details}
The CSCT model variants are trained for 150 epochs, with a learning rate of 1e-5, and a batch size of 128. The underlying transformer model consists of 2 encoder and 2 decoder layers, 8 attention heads, a fully connected layer size of 512, and an embedding size of 128. The dropout rate is set to 10 \% at both training and inference time. The loss function is defined  as a weighted combination of the lateral and longitudinal cross entropy errors, and the weights are learnable parameters fitted during training. 

\subsection{Prediction performance}
\begin{figure}
	\centering
	\begin{subfigure}{0.49\textwidth}
		\includegraphics[width=\textwidth]{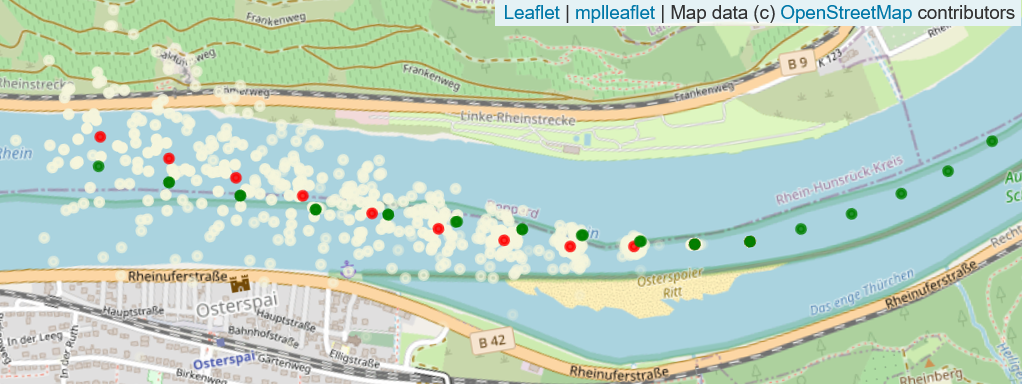}
		\caption{G-CSCT}
		\label{fig:ex1}
	\end{subfigure}
	\hfill
	\begin{subfigure}{0.49\textwidth}
		\includegraphics[width=\textwidth]{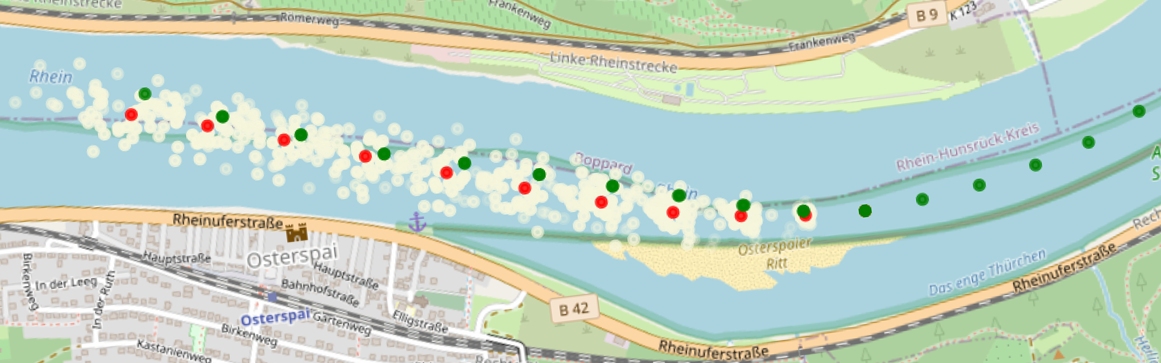}
	\caption{R-CSCT}
	\label{fig:ex2}
\end{subfigure}
	\begin{subfigure}{0.49\textwidth}
	\includegraphics[width=\textwidth]{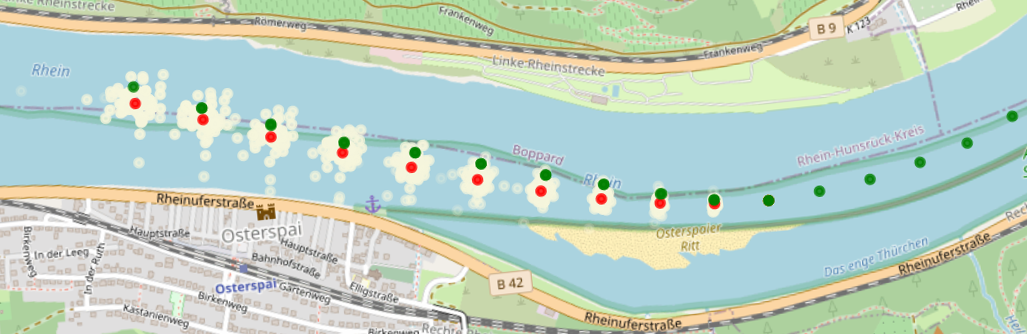}
\caption{N-CSCT}
\label{fig:ex3}
\vspace{-0.6em}
\end{subfigure}
\caption{Model predictions with MC dropout (beige), final trajectory prediction obtained from ($\overline{\hat{y}}_k$) (red), and ground truth (green). Opacity indicates distance from red positions.}\label{fig:ex}
\vspace{-1.5em}
\end{figure}

The absolute trajectory error (ATE) \cite{DijtPimandMettesPascal.2020} is used as error metric. It is defined as
\begin{equation}
	ATE = \sqrt{\frac{\sum_{t=T_{obs}+1}^{T_{obs}+n}\lVert \hat{p}_t-p_t \rVert^2 }{n}}, 
\label{formula:ate}
\end{equation}
where $p_t$ is the ground truth and $\hat{p}_t$ is the predicted position at time $t$ obtained by recursively calculating the predicted position from the current dislocation output $\hat{x}_t \in \lbrace \hat{x}_t^{glob}, \hat{x}_t^{riv},$ $\hat{x}_t^{nav}\rbrace$ and the last predicted position $\hat{p}_{t-1}$. 
Several ($N$) output sequences are generated by applying MC dropout. 
The trajectory obtained from the final prediction sequence $\overline{\hat{y}}_k$  (red dots in Fig. \ref{fig:ex}) is used to calculate the ATE for the test sample $k$. It is obtained by averaging the predicted dislocation components at each time step as
\begin{equation}
	\overline{\hat{y}}_k =  
	\begin{pmatrix}
		\frac{\sum_{i=1}^{N}\hat{x}^i_{T_{obs+1}} }{N}, \hdots , \frac{\sum_{i=1}^{N}\hat{x}^i_{T_{obs+n}} }{N} 
	\end{pmatrix}.
\end{equation}
In Table \ref{tab:ATE}
, the average ATE obtained on the test datasets by each model variant are shown for varying prediction horizons. The best performance for short-term predictions are obtained by G-CSCT which - apart from the curvature - does not have information of spatial limitations. A possible reason for the lower short-term performance of R-CSCT and N-CSCT might be the influence of the discrepancy between hectometers and true Euclidean distance which probably cannot completely be solved by the measures described in Section \ref{sec:ImprovedDislocationData}.  
For the first few minutes, G-CSCT can perform well as vessels have a high inertia and waterway shapes are typically smooth and do not change frequently during a vessel's trip of 10 min, especially in the upstream direction due to lower speeds. 
The longer the prediction window becomes, the more difficult the correct prediction of the vessel's behavior using the spatial-agnostic model will be. For more than 3 min, R-CSCT and N-CSCT perform better despite the mentioned KM and Euclidean distance discrepancy.  
The best model for long-term predictions is N-CSCT for both datasets and similar errors are obtained. For G-CSCT and R-CSCT, higher error differences between the Rhine and Danube models exist. The baseline takes both the mean deviation of the typical KM distance per minute and the average Euclidean distance from the typical route during $T_{obs}$ and extrapolates them into the future. This statistical baseline obtains competitive results  especially in the case of the Danube dataset, for which the lowest mean errors for medium-term predictions are obtained. Lower deviations of the typical behavior of vessels navigating on the Danube river can be expected as less traffic and thus less maneuvering necessity exists compared to the Rhine.
\begin{table}[]	
	\centering
	\caption{Absolute trajectory error with standard deviation}\label{tab:ATE}
	{\setlength\extrarowheight{2pt}
		\resizebox{\columnwidth}{!}{
			\begin{tabular}{ccccc}
				\multicolumn{5}{c}{\textbf{Danube}}                                                                                                                                        \\ \hline
				\multicolumn{1}{|c|}{\textbf{Time [min]}} & \multicolumn{1}{c|}{\textbf{G-CSCT}}       & \multicolumn{1}{c|}{\textbf{R-CSCT}} & \multicolumn{1}{c|}{\textbf{N-CSCT}}    & \multicolumn{1}{c|}{\textbf{Baseline}}     \\ \hline
				\multicolumn{1}{|c|}{\textbf{1}}         & \multicolumn{1}{c|}{\textbf{6.08$\pm$7.71}}   & \multicolumn{1}{c|}{9.73$\pm$8.55}      & \multicolumn{1}{c|}{8.18$\pm$ 8.08}     & 
				\multicolumn{1}{c|}{11.44$\pm$9.95}         \\ \hline

				\multicolumn{1}{|c|}{\textbf{3}}         & \multicolumn{1}{c|}{22.74$\pm$18.01}          & \multicolumn{1}{c|}{22.41$\pm$18.15}    & \multicolumn{1}{c|}{\textbf{16.69$\pm$16.97}} & 
				\multicolumn{1}{c|}{19.54$\pm$19.44} \\ \hline

				\multicolumn{1}{|c|}{\textbf{5}}         & \multicolumn{1}{c|}{43.48$\pm$ 29.79}         & \multicolumn{1}{c|}{37.75$\pm$30.05}    & \multicolumn{1}{c|}{\textbf{25.30$\pm$18.10}} & 
				
				\multicolumn{1}{c|}{28.41$\pm$30.92} \\ \hline

				\multicolumn{1}{|c|}{\textbf{10}}        & \multicolumn{1}{c|}{107.34$\pm$61.55}         & \multicolumn{1}{c|}{80.44$\pm$ 61.51}   & \multicolumn{1}{c|}{\textbf{48.02$\pm$51.11}} &

				\multicolumn{1}{c|}{52.36$\pm$61.08}  \\ \hline
				
				\multicolumn{5}{c}{\textbf{Rhine}}                                                                                                                                         \\ \hline
				\multicolumn{1}{|l|}{\textbf{Time [min]}} & \multicolumn{1}{c|}{\textbf{G-CSCT}}       & \multicolumn{1}{c|}{\textbf{R-CSCT}} & \multicolumn{1}{c|}{\textbf{N-CSCT}} & \multicolumn{1}{c|}{\textbf{Baseline}}        \\ \hline
				\multicolumn{1}{|c|}{\textbf{1}}         & \multicolumn{1}{c|}{\textbf{3.94$\pm$4.65}}   & \multicolumn{1}{c|}{8.71$\pm$6.87}     & \multicolumn{1}{c|}{8.47$\pm$7.02}     & \multicolumn{1}{c|}{16.26$\pm$14.27}             \\ \hline
				\multicolumn{1}{|c|}{\textbf{3}}         & \multicolumn{1}{c|}{\textbf{16.03$\pm$11.74}} & \multicolumn{1}{c|}{17.68$\pm$13.09}   & \multicolumn{1}{c|}{16.46$\pm$13.35}     & \multicolumn{1}{c|}{22.45$\pm$17.56}           \\ \hline
				\multicolumn{1}{|c|}{\textbf{5}}         & \multicolumn{1}{c|}{30.53$\pm$20.18}          & \multicolumn{1}{c|}{26.56$\pm$19.66}    & \multicolumn{1}{c|}{\textbf{23.55$\pm$19.66}}    & \multicolumn{1}{c|}{29.06$\pm$23.03} \\ \hline
				\multicolumn{1}{|c|}{\textbf{10}}        & \multicolumn{1}{c|}{93.88$\pm$49.46}          & \multicolumn{1}{c|}{52.46$\pm$40.30}    & \multicolumn{1}{c|}{\textbf{41.17$\pm$37.11}}    & \multicolumn{1}{c|}{46.89$\pm$41.13}  \\ \hline
			\end{tabular}
	}}
\vspace{-0.5em}
\end{table}

\subsection{Uncertainty analysis}
\begin{figure}
	\vspace{1em}
	\centering
	\includegraphics[width=0.49\textwidth]{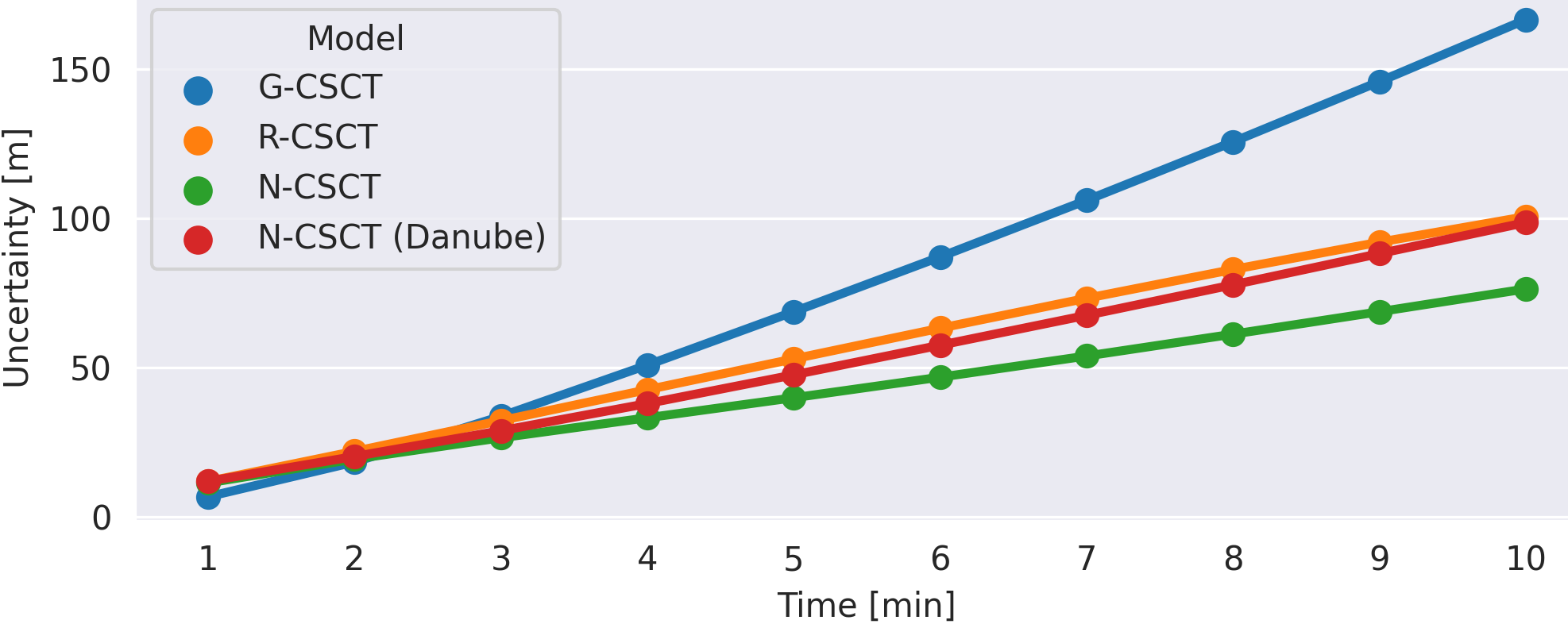}
\caption{Average uncertainty per predicted time step}\label{fig:unc_time}
\vspace{-1em}
\end{figure}
\begin{figure}
	\centering
	\includegraphics[width=0.49\textwidth]{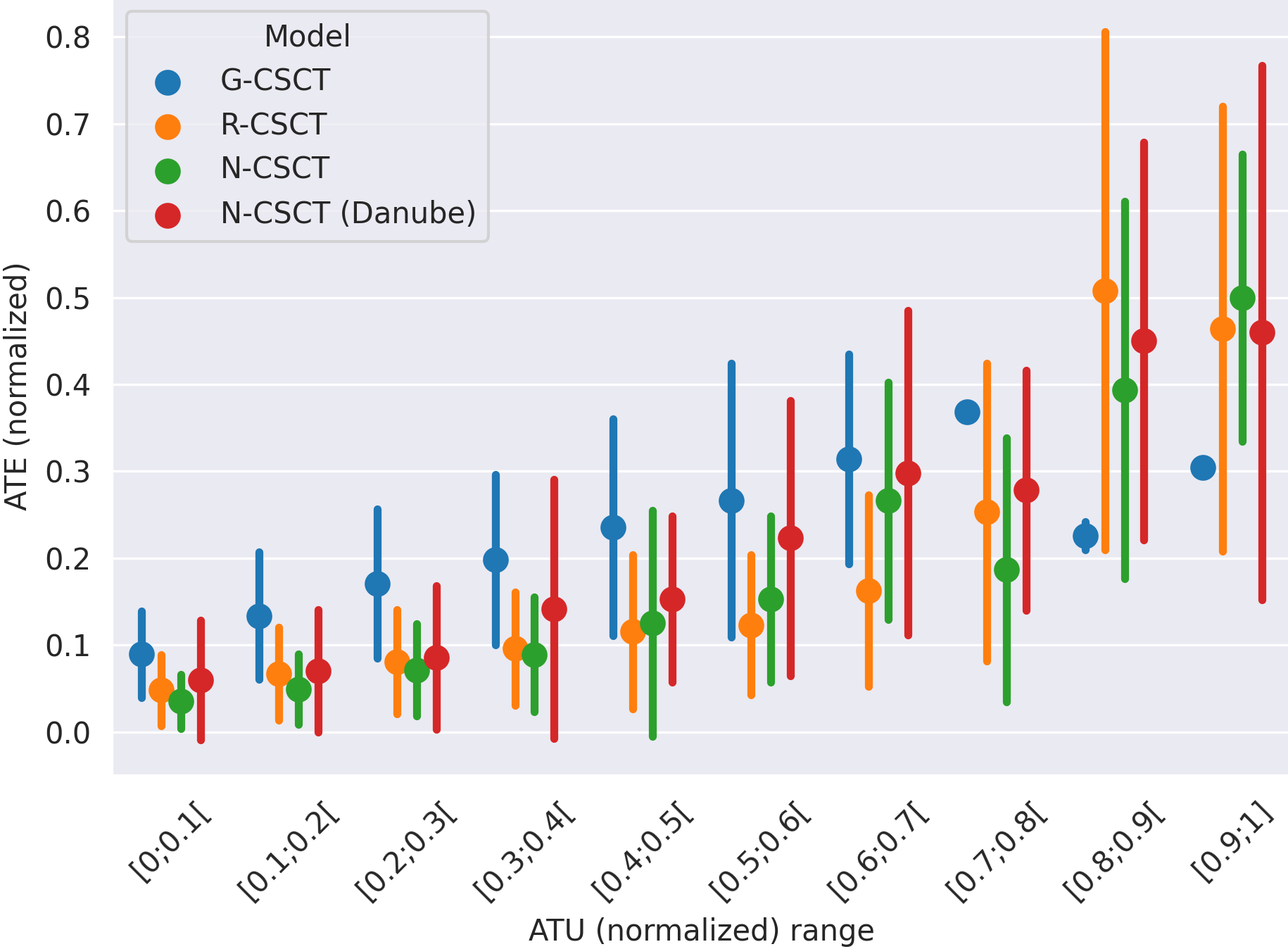}
	\caption{Relationship between absolute trajectory error (ATE) for  n=10 and absolute trajectory uncertainty (ATU)}\label{fig:error_unc}
\vspace{-1.2em}
\end{figure}

The uncertainty at predicted time step $t$ for a given input sequence is calculated by the Euclidean norm 
\begin{equation}
u_t := \biggl \lVert \begin{pmatrix}\sigma_t^{east}\\\sigma_t^{north}\end{pmatrix} \biggl \rVert, 
\end{equation} 
with
\begin{equation*}
\sigma_t^{east} = \sqrt{\frac{1}{N}\sum_{i=1}^{N}\bigl(\hat{e}_{t}^{i}-\overline{\hat{e}}_t\bigl)^2}, 
\end{equation*}
where $\hat{e}_t^i$ is the easting coordinate of the position $\hat{p}^i_t$ obtained from the $i$th prediction, and $\overline{\hat{e}}_t$ is the mean easting value of the $N$ predictions for time step $t$ obtained for the input sequence using MC dropout. The definition of $\sigma_t^{north}$ is analogous to  $\sigma_t^{east}$. 
In  Fig. \ref{fig:unc_time}, the average uncertainty per predicted time step obtained on the test datasets is depicted. For the Danube case, only the results of N-CSCT are shown for better visibility. The increasing uncertainty over the prediction horizon is also visible in the examples of Fig. \ref{fig:ex} by the spread of the predictions (beige). The G-CSCT illustrated in Fig. \ref{fig:ex1} appears as least and the N-CSCT illustrated in Fig. \ref{fig:ex3} as most confident for the given input sequence. 
The increasing uncertainty over time shows that the models are able to provide information on their reliability. In well-calibrated models, uncertainty and prediction error are positively correlated \cite{krishnan2020improving}. To further analyze the relationship between prediction error and uncertainty, the ATEs of the final predicted trajectories are grouped by the absolute trajectory uncertainty (ATU) defined here similarly to the ATE (Eq. \ref{formula:ate}) by replacing $\lVert \hat{p}_t-p_t \rVert$ with $u_t$. To compare the model variants' uncertainty-error relationship without the influence of the error and uncertainty magnitude, both ATE and ATU are normalized. The mean and standard deviation of the normalized ATEs per 0.1 normalized ATU interval are shown in Fig. \ref{fig:error_unc}. All model variants show a positive ATE-ATU correlation in the beginning. The highest uncertainties are, however, not necessarily attributed to the highest ATEs. Thus, further effort is required to improve the models and reach a better calibration. The mean and standard deviation of the Danube N-CSCT ATE per ATU interval are higher than those of the Rhine N-CSCT most of the times.  
This can be caused by the smaller amount of training data.  

\section{CONCLUSIONS \& OUTLOOK}
The separate consideration of spatial context and dislocation information in DL-based inland vessel trajectory prediction models is replaced by defining the dislocation in relation to the spatial context directly. A river- and a navigation-specific reference system are proposed. Both make use of the waterway kilometerization and mitigation strategies to address the discrepancy between waterway kilometer and Euclidean distance.  
The models trained with river- or navigation-specific dislocation information do not rely on an increased architecture complexity for the context consideration. 
Compared to their spatial-agnostic counterpart, better medium and long-term predictions are obtained. 
Relying on statistics and extrapolation methods can already provide good prediction results for the given data. The model which learns to predict deviations of the typical behavior obtained from statistics is able to outperform the statistical baseline. 
In future works, delegating the task of learning the deviation of typical behaviors will therefore be preferred. 
The epistemic uncertainty is assessed by using Monte Carlo dropout. A  positive correlation between prediction horizon length and model uncertainty is observed. The models' error for a prediction horizon of 10 min is in average positively correlated with the proposed uncertainty measure for low and medium uncertainties. Attributing the highest uncertainty correctly remains difficult. 
Further possible extensions of the proposed approach include the enabling of directly processing irregularly-sampled instead of interpolated AIS data typically used for sequence-to-sequence vessel trajectory prediction models and the consideration of the uncertainty in the data.  
As interaction is probably the main reason of deviation from typical behaviors as indicated by the performance difference of the baseline on the two datasets, the inclusion of the surrounding traffic will be considered. 

\addtolength{\textheight}{-12cm}   

\section*{ACKNOWLEDGMENT}

The authors thank the BAW for the provision of AIS and river-specific data, the training infrastructure, and especially Michael Schröder, Christian Noß, and Johannes Schwarz-Beutel for valuable discussions and ideas. 

\bibliographystyle{IEEEtran}
\bibliography{bibliography}

\end{document}